\documentclass[
twocolumn,
]{ceurart}

\usepackage{listings}
\usepackage{caption}
\usepackage{graphicx}
\PassOptionsToPackage{table}{xcolor}
\begin{document}

\copyrightyear{2023}
\copyrightclause{Copyright for this paper by its authors.
  Use permitted under Creative Commons License Attribution 4.0
  International (CC BY 4.0).}

\conference{Proceedings of the Sixth Workshop on Automated Semantic Analysis of Information in Legal Text (ASAIL 2023), June 23, 2023, Braga, Portugal.}

\title{Argumentative Segmentation Enhancement for Legal Summarization}


\author[1,3]{Huihui Xu}[%
email=huihui.xu@pitt.edu
]
\cormark[1]
\address[1]{Intelligent Systems Program, University of Pittsburgh, PA, USA}

\author[1,2,3]{Kevin Ashley}[%
email=ashley@pitt.edu
]
\address[2]{School of Law, University of Pittsburgh, PA, USA}

\address[3]{Learning Research and Development Center, University of Pittsburgh, PA, USA}
\cortext[1]{Corresponding author.}

\begin{abstract}
We use the combination of argumentative zoning \cite{teufel2009towards} and a legal argumentative scheme to create legal argumentative segments. Based on the argumentative segmentation, we propose a novel task of classifying argumentative segments of legal case decisions. GPT-3.5 is used to generate summaries based on argumentative segments.  In terms of automatic evaluation metrics, our  method generates higher quality argumentative summaries while leaving out less relevant context as compared to GPT-4 and  non-GPT models. 
\end{abstract}


\maketitle

\section{Introduction}

Automatic text summarization is a process of automatically generating shorter texts that convey important information in the original documents \cite{radev2002introduction}. There are in general two different approaches for automatic text summarization: extractive summarization and abstractive summarization \cite{el2021automatic}. Extractive summarization can be conceptualized as a sentence classification task, where the algorithm selects important sentences from the original document directly \cite{widyassari2020review}.  Abstractive summarization  can be a more  natural way of summarizing in terms of novel words and expressions \cite{gupta2019abstractive}. Authors of \cite{zhong-litman-2022-computing, dong2021discourse} have experimented with several extractive summarization methods in  domains like law and science. 

Abstractive summarization is flourishing in recent years because of the rise of large pre-trained language models, like BART \cite{lewis2020bart}, T5 \cite{raffel2020exploring}, and Longformer \cite{beltagy2020longformer}. However, those models still require sizable training datasets to tackle a new task. For example, a language model trained on a Wikipedia text corpus requires fine-tuning on a legal dataset. In addition, unlike news articles, legal case decisions are longer and contain argumentative structures \cite{ elaraby2022arglegalsumm}. 
While some summarization approaches are beginning to take the argumentative structure of a legal case decision into account (e.g., \cite{elaraby2022arglegalsumm}), none do so in a zero-shot setting. 

In this paper, we conduct a study of summarizing  argumentative segments extracted from a legal document using the latest GPT-3.5 model (text-davinci-003) and GPT-4 \cite{openai2023gpt4} model. The new GPT-3.5 version is based on  InstructGPT \cite{ouyang2022training}, which is also trained with reinforcement learning from human feedback (RLHF). Despite its potential for generating high quality summaries, the GPT-3.5 model has a 4,097-token input limitation. This is a disadvantage for summarizing long legal documents. Our work employs a method of cutting the long documents into shorter segments while still preserving argumentative components. GPT-4 is also trained with RLHF like GPT-3.5 but with more capability. For example, GPT-4 can handle 8,192\footnote{There is another version of the model that supports 32,768 tokens.} tokens as input, which has doubled GPT-3.5's input length. Even though GPT-4 can handle longer documents, there are still some legal documents that exceed the input limitation. Besides, we believe that legal argument mining and argumentative zoning can extract argumentative segments that can help models to generate better legal summaries. 

In order to extract argumentative segments from legal decisions, we propose a novel task for automatically classifying segments as argumentative or non-argumentative segments. This task stems from Argumentative Zoning (AZ) addressed in \cite{teufel1999argumentative, teufel2009towards}. Teufel et al. define the task of AZ as a sentence level classification with mutually exclusive categories given an annotation scheme. AZ divides a paper into zones on the basis of the content knowledge claim in the corresponding segment \cite{teufel2009towards}. We adopt the reasoning behind AZ and divide textual segments into argumentative or non-argumentative segments by examining if any argumentative sentences exist in the corresponding segment. The identified argumentative segments are then fed into the model for generating summaries. 

\begin{figure*}[h!]
    \includegraphics[width=0.9\textwidth]{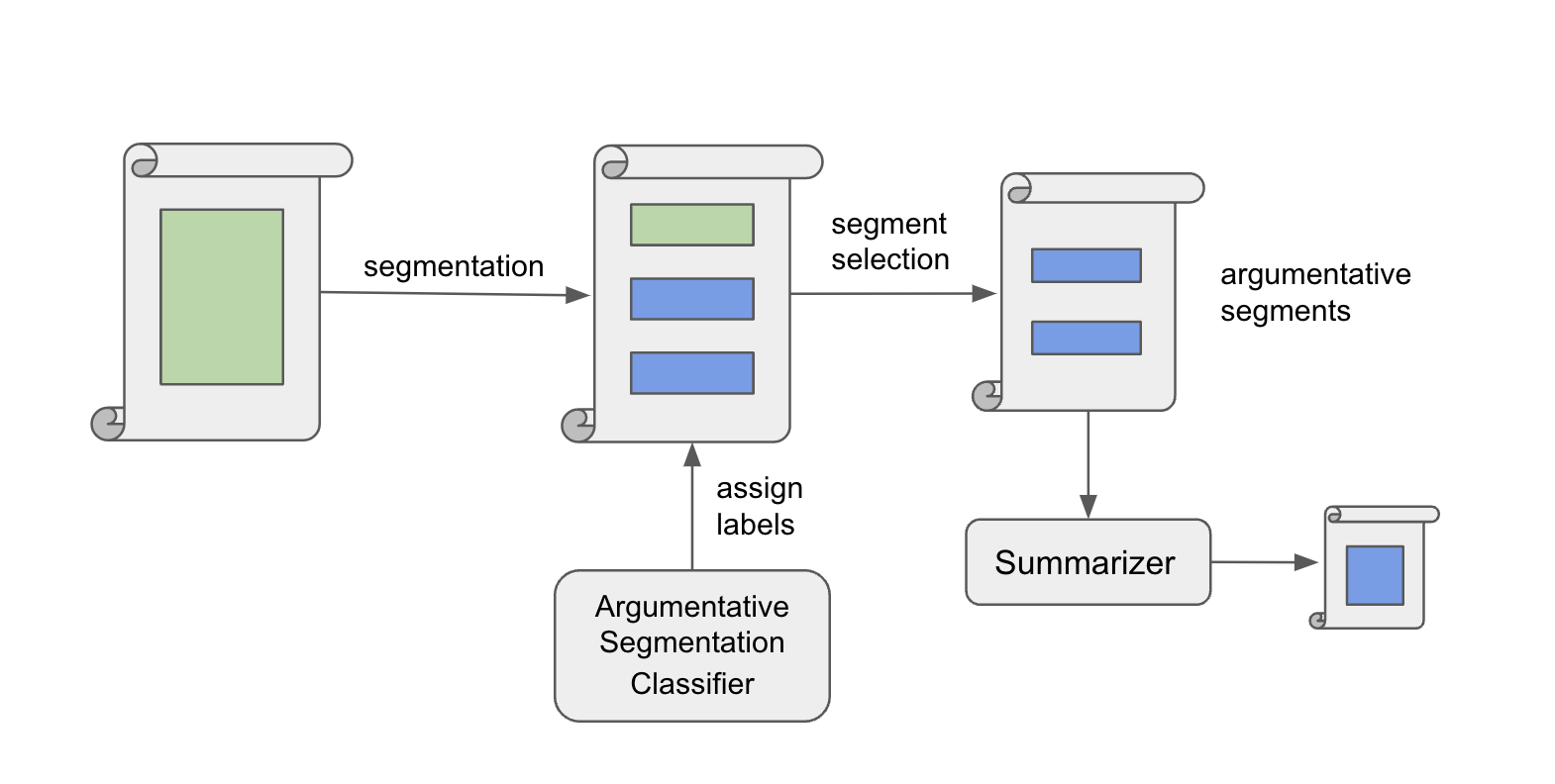}
    \centering
    \caption{Pipeline of generating argumentative segmentation enhanced legal summarization.}
    \label{fig:pipeline}
\end{figure*}

Figure \ref{fig:pipeline} illustrates the summarization pipeline of our approach. The pipeline  comprises three stages.  First, the document, a full-text legal opinion is segmented into several parts. Then, every segment is assigned a label based on the existence of argumentative sentences using a classifier. Finally, the predicted argumentative segments are fed into the model. The model will summarize each segment and concatenate them as the final summary for the decision. 

Our contributions in this work are: (1) We propose a novel task of predicting argumentative segments in the legal context. (2) We show that our approach for using argumentative segments to guide summarizing is effective. (3) We overcome the token limitation of  GPT-3.5 when applied to long document summarization and show a promising result in a zero-shot setting.  

\section{Related Work}
\subsection{Argumentative Zoning}
Teufel, et al. \cite{teufel1999argumentative} first proposed and defined the task of AZ as a sentence level classification with mutually exclusively categories given a certain annotation scheme for scientific papers. The earliest scheme includes seven categories of zones, such as Aim and Background. The annotation scheme is based on the rhetoric roles employed by authors. For example, one can identify sections that cover the background of the scientific research in a technical paper among other sections. Later, \cite{teufel2009towards} made attempts toward discipline-independent argumentative zoning in two different domains. The idea of AZ is seeking to extract the structure of research components that follows authors' knowledge claims. As a result, there are different AZ schemes for different domains, such as \cite{liakata2010corpora} for chemistry research articles, \cite{dayrell2012rhetorical} for physical sciences and engineering and life and health sciences. AZ was later adopted for legal documents in \cite{hachey2005automatic, moens1997automatic}. Since AZ classified sentences into different categories, it is helpful for generating summaries for long documents. \cite{el2022platform} proposed a tool for AZ annotation and summarization. However, AZ annotation for legal documents can be expensive. We propose to leverage our sentence level annotation for AZ in the context of argumentative segmentation classification. 

\subsection{Legal Argument Mining}
Legal Argument mining aims to extract legal argumentative components from legal documents. Most argument mining work consists of three sub-tasks: identifying argumentative units, classifying the roles of the argumentative units, and detecting the relationship between the argumentative units.  \cite{mochales2008study} explored the argumentative characteristics of legal documents.\cite{saravanan2010identification, feng2011classifying} 
identified  rhetorical roles that sentences play in a legal context. Early work in legal argument mining rely on word patterns and syntactic features \cite{mochales2011argumentation, palau2009argumentation}. Recently, contextual embedding has been used for legal argument mining \cite{xu2021toward, xu2021accounting}, 
like Sentence-BERT \cite{reimers2019sentence} and LegalBERT \cite{zheng2021does} embedding. 
\cite{xu2021toward, xu2021accounting} have proposed a legal argument triples scheme to classify sentences for summarizing legal opinions in terms of Issues, Reasons, and Conclusions. 

\subsection{Summarization Methods and GPTs}
As noted, the automatic summarization methods can be categorized as extractive or abstractive. 
Most ML approaches for learning to extract  sentences for summarizing documents  
are unsupervised \cite{yin2015optimizing, hirao2013single}. They are  based on learning sentence importance scores for selecting sentences to form summaries. The development of better sentence representations, like Sentence-BERT, has lead to  improvements in generating better summaries \cite{miller2019leveraging}. 

Recent research applying sequence-to-sequence neural models to summarization is gaining more attention. \cite{see2017get} proposed a pointer generator architecture for generating higher quality abstractive summaries. Transformer-based sequence-to-sequence models, like BART (Bidirectional and Auto-Regressive Transformer), T5 (Text-to-Text Transfer Transformer) and Longformer, have been used in generating abstractive summaries. \cite{elaraby2022arglegalsumm} incorporate legal argumentative structures into sequence-to-sequence model to further enhance the quality of summaries. In this work, Longformer Encoder-Decoder (LED), T5 and BART serve as the baseline for our experiments. 

The mainstream transformer-based models, however, require a curated training set to adapt to a new domain. The success of prompt-based models provides a new way of solving the domain adaption problem by learning from a large unlabelled dataset. GPT-3.5 and GPT-4, developed by OpenAI, are the examples of prompt-based models. 
\cite{goyal2022news} investigated  how  zero-shot learning with GPT-3 compares with fine-tuned models on news summarization task. Their results show that GPT-3 summaries are preferred by humans. Our work focuses instead on legal summarization and takes argumentative structure into account. The results show a higher performance in terms of automatic evaluation metrics by taking account of argument structures. We further experimented with GPT-4 on legal summarization, since it has a larger context window compared to GPT-3.5. Our findings demonstrate that considering the argumentative structure leads to improved summaries.

\section{Legal Decision Summarization Dataset}

We use the legal decision summarization dataset provided by the Canadian Legal Information Institute (CanLII).\footnote{\url{https://www.canlii.org/en/}} The summaries are prepared by attorneys, members of legal societies, or law students. The basic statistics of the annotated dataset are listed in Table \ref{tab:stats}. The court decisions involve a wide variety of legal claims.
The average length of the court decisions is 4,382 tokens. It exceeds the token limitation of GPT-3.5 (4,097 tokens). This motivates us to explore argumentative segmentation to reduce the input document length. 

\begin{table*}[h]
  \caption{Statistics of CanLII annotated court decision and summary pairs. }
  \label{tab:stats}
  \begin{tabular}{cccc}
    \toprule
    \textbf{Type} & \textbf{Avg. \# of tokens} & \textbf{Max. \# of tokens} & \textbf{Min. \# of tokens}\\
    Court decision & 4382 &62785 & 122\\
    Human-written summary & 273 &2072 &17\\
   
    \bottomrule
  \end{tabular}
\end{table*}
In prior work, researchers conceptualized a type system for annotating sentences in legal case decisions and summaries, which includes: 
\textbf{Issue} -- Legal question which a court addressed in the case;
 \textbf{Conclusion} --  Court's decision for the corresponding issue;
\textbf{Reason} -- Sentences that elaborate on why the court reached the Conclusion \cite{xu2020using}. 
Those sentences are referred to as IRC triples. We have accumulated 1,049 annotated legal case decision and summary pairs. \cite{elaraby2022arglegalsumm, zhong-litman-2022-computing} use the same dataset for legal summarization tasks. \cite{elaraby2022arglegalsumm} use the IRC annotations as markers to inform models with argumentative information. \cite{zhong-litman-2022-computing} explored the structure of legal decisions and used the annotated dataset as the basis for domain-specific evaluation of summaries. 

In this work, we use the idea of argumentative zoning to further expand the use of IRC triples. The documents in the dataset have already been split at a sentence level. They have not yet been split into paragraphs or annotated in terms of explicit rhetorical zones. We adopt C99 \cite{choi2000advances}, a domain-independent linear text segmentation algorithm, to further segment the legal case decisions on a higher level. This algorithm measures the similarity between all sentence pairs to generate a similarity matrix. The similarity between a pair of sentences $x$, $y$ is calculated using cosine similarity. Sentence-BERT is used for representing all sentences in the same space before computing the similarity scores. Then we cluster the neighboring sentences into groups based on the similarity scores.



 Here, we propose a novel task -- argumentative segmentation classification. For each group of sentences, we assign an ``argumentative segment (1)'' if there exists one or more  IRC sentences, or a ``non-argumentative segment (0)'' otherwise. This combines the idea of argumentative zoning with semantic segmentation. Table \ref{tab:arg_example} shows an example of an argumentative segment. As the example shows,  segment no. 9 is labeled as an argumentative segment because of the existence of a conclusion sentence.

We split our data into 80\% training, 10\% validation and 10\% test datasets. 

\begin{table*}
    \caption{An example of segmented legal case decision from the annotated dataset. The text with \colorbox{lightgray}{gray} background is an annotated conclusion. Since it appears in segment no.9, the segment is labelled as argumentative. }
    \label{tab:arg_example}
    \begin{tabular}{|p{0.7in}|p{4.3in}|}
    \hline
    ..&.. \\
    \hline
      \textbf{segment no.8 (non-argumentative segment)} & III As matter of public policy, the Crown is not required to disclose the name of the confidential informer. If the Information discloses too much information about the informer and his means of knowledge, the identity of the informer will become apparent. As result, the Crown has to take refuge in the kind of language employed in this Information.\\
    \hline
    \textbf{segment no.9 (argumentative segment)} & note the type of language used by the peace officer has been accepted, as compliance with the section, in other cases: see Re Lubell and The Queen (1973), 1973 CanLII 1488 (ON SC), 11 C.C.C. (2d) 188 (Ont. H.C.); Re Dodge and The Queen (1985), 1984 CanLII 59 (NL SC), 16 C.C.C. (3d) 385 (Nfl. S.C.). Perhaps more information could have been provided, however, there was information upon which the respondent, acting judicially, could be satisfied that search warrant should issue. Courts should not be too technical when scrutinizing the Information in support of search warrant; \colorbox{lightgray}{substantial compliance with s. 443 is sufficient.}\\
    \hline
  
  \end{tabular}
\end{table*}

\section{Experiments and Results}

\subsection{Argumentative Segment Classification}
\begin{figure}[h!]
    \centering
    \includegraphics[scale=0.45]{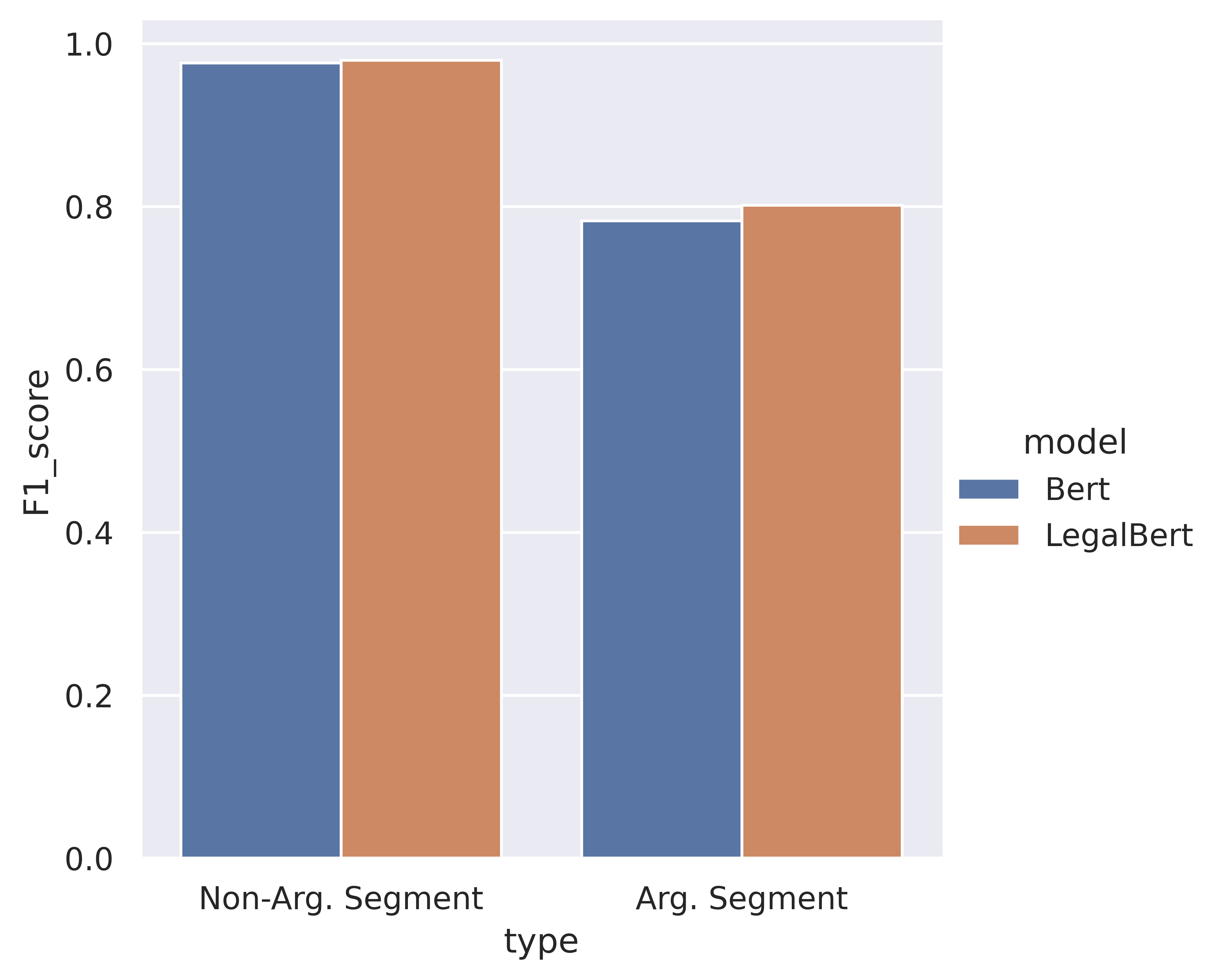}
    \caption{Argumentative segmentation prediction results. }
    \label{fig:seg_classification}
\end{figure}

Every legal case decision in our dataset has been split into segments using the C99 algorithm. Table \ref{tab:c99_seg} shows the results of C99 segmentation. From the table, the average number of argumentative segments is 6 in a legal decision while the number of non-argumentative segments is 59. Thus, the number of argumentative segments is much less than non-argumentative segments in legal decisions. We performed a segment-level classification using the mentioned data split. We conducted experiments with different transformer models, BERT \cite{devlin-etal-2019-bert} and LegalBERT\cite{zheng2021does}. We use those models to predict the argumentativeness of segments (i.e., argumentative segment, or non-argumentative segment). Figure \ref{fig:seg_classification} shows the results of the binary classification. The figure shows LegalBERT achieved a better classification result compared to BERT. LegalBERT achieved 80.14\% $F1$ score while BERT has 78.24\%. As a result, we chose to use LegalBERT's predictions to select input segments for the following summarization task. 
\begin{table*}[h!]
  \caption{Statistics of text segmentation on court decisions using C99 with Sentence-BERT('bert-base-nli-stsb-mean-tokens') embedding. }
  \label{tab:c99_seg}
  \begin{tabular}{lccc}
    \toprule
    \textbf{Type} & \textbf{Avg. \# of segments} & \textbf{Max. \# of segments} & \textbf{Min. \# of segments}\\
    Argumentative segmentation & 6 & 30 & 1\\
    Non-argumentative segmentation & 59&732 &0\\
    \hline
    Total &65 &737 &2 \\
    \bottomrule
  \end{tabular}
\end{table*}

\subsection{Baselines}
We use two different types of baselines for our proposed argumentative segmentation enhanced summarization method. One is non-GPT abstractive summarzation model, like LED, T5, and BART. The other one is vanilla GPT-3.5 and GPT-4. They are both developed by OpenAI. 

The GPT-3.5 model is an auto-regressive language model. This model can generate high quality news summaries in a zero-shot setting according to \cite{goyal2022news}.  We used the latest version, text-davinci-003, in our work just released in November 2022. There is little or no work, however,  measuring how well the model performs on legal documents. GPT-4 is a multi-modal large language model, which is more capable than GPT-3.5. GPT-4 was released in March 2023, and it is by far the most advanced large language model in the field.  

\subsection{Prompting for GPT-3.5 and GPT-4}
As mentioned, GPT-3.5 and GPT-4 are both prompt-based model. In order to use GPT-3.5 and GPT-4 to summarize a chunk of text, we have to inform the model of the type of task to perform. In our experiment, we add a short text ``TL;DR'' immediately after the input text. ``TL;DR'' is an abbreviation for ``Too Long; Don't Read'', and \textbackslash n is the change of a new line. ``TL;DR" instructs GPT-3.5 and GPT-4 to summarize the text in a fewer number of words. The example prompt is listed below: 
\begin{equation}
     \{\{Text\}\} + \backslash n TL;DR 
\end{equation}
We only need to control the max output tokens and temperature without fine-tuning on our dataset. This is a zero-shot setting because the model does not see any human-written summaries before generating summaries. We noticed that the lengths of generated
summaries are consistent. The average lengths of model-generated summaries are reported in Table \ref{tab:summ_results}, Table \ref{tab:gpt3.5_results} and Table \ref{tab:gpt4_results}. 

For the baseline GPT-3.5 model, we chunk the original document into lengths which the model accepts. We tried different lengths, and finally settled on 2,500 tokens to avoid an ``over token request limitation error.'' The argumentative segmentation enhanced GPT-3.5 model does not have this problem because the argumentative segments are  shorter than GPT-3.5's token limitation. It also helps GPT-3.5 to focus on the chunks of text that have important argument-related information. Even though GPT-4 has much longer context length, it still falls short for dealing with some long documents. We set 7,500 tokens as the limit of prompt length to avoid ``over token request limitation error.''

\subsection{Results}
Rouge-1, Rouge-2, Rouge-L, BLEU, METEOR, and BERTScore are used to measure the performance. 
Rouge stands for Recall-oriented Understudy for Gisting Evaluation \cite{lin2004rouge}. Rouge-based evaluation metrics examine lexical overlap between generated and reference summaries. BLEU stands for Bilingual Evaluation Understudy \cite{papineni2002bleu}; it measures word overlap taking order into account. It is often used to measure the quality of machine translation. METEOR \cite{banerjee2005meteor} computes the similarity between generated and reference sentences by mapping unigrams. BERTScore \cite{Zhang*2020BERTScore:} uses contextual token embedding to compute similarity scores between generated and reference summaries on a token level. 

Table \ref{tab:summ_results} shows the test set results of different summarization models in different experimental settings. We first experimented with those non-GPT models in a zero-shot setting, and the results are shown in parentheses. Since zero-shot performance is not good, we further fine-tuned those models on the training set. We adopt some of the training hyperparameters from \cite{elaraby2022arglegalsumm}: initializing LED and BART with learning rate of $2e^{-5}$, T5 with learning rate of $1e^{-4}$; and training both models for 10 epochs; set maximum input length is 6144 words for LED and T5 and 1024 for BART; maximum output length is set to 512 tokens for all the models. LED, T5 and BART outperform baseline GPT-3.5 and GPT-4 in term of automatic evaluation metrics. We also find that LED, T5 and BART produce longer summaries than GPT-3.5 and GPT-4 on average, which might directly contribute to the higher scores across some of the metrics.

\begin{table*}[h]
  \caption{Summarization results on the test set. The numbers within  parentheses represent zero-shot outcomes of the models; the numbers with no parentheses are results obtained after fine-tuning on the training set. Bolded numbers are the best results. We use GPT-3.5 (``text-davinci-003") for the experiments. More detailed experiment results are shown in Table \ref{tab:gpt3.5_results}. ``*'' represents the best result from Table \ref{tab:gpt4_results}. ``**'' represents the best result from the sets of experiments from Table \ref{tab:gpt3.5_results}.} 
  \label{tab:summ_results}
  \resizebox{\textwidth}{!}{
  \begin{tabular}{clccccccc}
    \toprule
    \textbf{Experiment} & \textbf{Model} &\textbf{Avg. length}& \textbf{Rouge-1} & \textbf{Rouge-2}  & \textbf{Rouge-L} &\textbf{BLEU} &\textbf{METEOR}& \textbf{BERTScore}\\
    \midrule

    Baseline& LED &393(387) &46.14(37.08) &21.84(12.33) &42.99(33.36) &14.28(5.95) &0.38(0.28) &85.98(81.49)\\
    Baseline& T5 &167(65) &45.30(21.39) &22.51(5.10) &42.11(19.47) &14.91(1.83) &0.28(0.10) &86.84(82.84)\\
    Baseline& BART-large &365(286) &44.26(27.56) &18.52(8.89) &37.53(24.94) &10.82(3.80) &0.32(0.18) &84.85(76.55)\\
    No Arg Seg. &GPT-3.5 &126 &40.88 &18.90 &37.63 &10.47 &0.23&86.56\\
    No Arg Seg. &GPT-4* &144 &43.46 &19.84 &30.17 &11.13 &0.26 &86.51\\
    Arg Seg. &GPT-3.5** &205 &\textbf{49.42} &\textbf{23.98} &\textbf{46.07} &\textbf{17.54} &\textbf{0.32} &\textbf{87.30}\\
    \bottomrule
  \end{tabular}
  }
\end{table*}

\begin{table*}[h!]
  \caption{GPT-3.5 model with different combinations of temperatures and max output tokens (temperature, max tokens). }
  \label{tab:gpt3.5_results}
  \resizebox{\textwidth}{!}{
  \begin{tabular}{ccccccccc}
    \toprule
    \textbf{Parameters} & \textbf{Avg. summary length  }&\textbf{Rouge-1} & \textbf{Rouge-2}  & \textbf{Rouge-L} &\textbf{BLEU} &\textbf{METEOR}& \textbf{BERTScore}\\
    \midrule
        (0, 32)  &128 &41.60 &19.05 &38.76 &10.88 &0.23 &86.02\\
        (0.3, 32) &129 &41.79 &19.07 &38.58 &10.87 &0.23 &86.05\\
        (0.5, 32) &129 &41.55 &18.94 &38.47 &10.38 &0.23 &86.06\\
        (0.8, 32) &128 &40.79 &17.85 &37.58 &9.34 &0.22 &85.92\\
        \hline
        (0, 64) &184 &47.78 &22.95 &44.51 &15.77 &0.30 &87.03\\
        (0.3, 64) &186 &47.79 &22.92 &44.47 &15.80 &0.30&86.98\\
        (0.5, 64) &183 &46.93 &21.80 &43.50 &14.55 &0.29 &87.00\\
        (0.8, 64) &183 &46.63 &21.29 &43.03 &13.79 &0.29 &86.93\\
        \hline
        (0, 128) &205 &49.42 &23.98 &46.07 &17.54 &0.32 &87.30\\         
        (0.3, 128) &203 &49.32 &23.72 &45.84 &16.77 &0.32 &87.32\\  
        (0.5, 128) &200 &48.84 &23.29 &45.23 &16.22 &0.32 &87.25\\
        (0.8, 128) &197 &47.22 &21.11 &43.39 &14.30&0.30&87.05\\
    \bottomrule
    \rowcolor{gray!40}
        \multicolumn{2}{c}{Average} &45.81 &21.33 &42.45 &13.85 &0.28&86.74\\
    \bottomrule
  \end{tabular}
}
\end{table*}

\begin{table*}[h!]

  \caption{GPT-4 model with different combinations of temperatures and max output tokens (temperature, max tokens). }
  \label{tab:gpt4_results}
 
  \begin{tabular}{ccccccccc}
    \toprule
    \textbf{Parameters} & \textbf{Avg. summary length  }&\textbf{Rouge-1} & \textbf{Rouge-2}  & \textbf{Rouge-L} &\textbf{BLEU} &\textbf{METEOR}& \textbf{BERTScore}\\
    \midrule
        (0, 128)  &106 &39.21 &17.52 & 27.33&8.33 &0.22 &86.18\\
        (0.3, 128) &109 &40.57 &18.41 &28.38 &8.66 &0.23 &86.30\\
        (0.5, 128) &106 &39.16 &17.80 &27.45 &8.71 &0.22 &86.26\\
        (0.8, 128) &110 &39.00 &16.75 &27.25 &8.07 &0.22 &86.18\\
        \hline
        (0, 256) &132 &42.76 &19.63 &29.80 &10.69 &0.26 &86.45\\
        (0.3, 256) &134 &42.64 &19.21 &29.36 &10.10 &0.26 &86.43\\
        (0.5, 256) &134 &43.42 &19.91 &30.35 &10.62 &0.26 &86.54\\
        (0.8, 256) &135 &42.58 &18.84 &29.06 &9.57 &0.26 &86.39\\
        \hline
        (0, 512) &144 &43.46 &19.84 &30.17 &11.13 &0.26 &86.51\\         
        (0.3, 512) &137 &43.07 &19.58 &29.96 &10.43 &0.26 &86.55\\  
        (0.5, 512) &142 &43.19 &19.48 &30.18 &10.12 &0.26 &86.49\\
        (0.8, 512) &140 &42.15 &18.50 &29.38 &9.99 &0.25 &86.45\\
    \bottomrule
    \rowcolor{gray!40}
     \multicolumn{2}{c}{Average} &41.77 &18.79 &29.06 &9.70 &0.25 &86.39\\
    \bottomrule
  \end{tabular}
\end{table*}

Table \ref{tab:gpt3.5_results} shows different combinations of two important control parameters in GPT-3.5: \verb|temperature| and \verb|max_tokens|. According to the official website,\footnote{\url{https://platform.openai.com/playground/p/default-tldr-summary?model=text-davinci-003}} \verb|temperature| ranges between 0 and 1 and controls the randomness of generated text. With a 0 \verb|temperature|, GPT-3.5 will select the most deterministic response, while a 1 \verb|temperature| is the most random. \verb|max_tokens| parameter controls the number of generated tokens. We found that the model generally performs better at a lower \verb|temperature|. For example, when the \verb|max_tokens| parameter is fixed at 128, the Rouge and BLEU scores decrease when the \verb|temperature| rises from 0 to 0.8. We also notice that the \verb|max_tokens| also affect the performance: when the \verb|temperature| is set to 0, the model with 128 \verb|max_tokens| achieves the best scores across all metrics except the BERTScore. We control GPT-4 with the same parameters, and the results are presented in Table \ref{tab:gpt3.5_results}.

Table \ref{tab:summary_example} shows the comparison between a reference summary and GPT generated summary when the input does not exceed either the GPT-3.5 and GPT-4 token limitations. We observe that the generated summaries provide similar information regarding the case facts. However, the argumentative segmentation enhanced GPT-3 generated summary provides additional information about the judge's considerations.



\begin{table*}[h!]
    \caption{Argumentative segmentation enhanced GPT-3.5 and baseline GPT-4 generated summaries versus reference summary.\colorbox{pink}{<Issue> </Issue>}, \colorbox{cyan}{<Reason> </Reason>} and \colorbox{lime}{<Conclusion> </Conclusion>} are markers for respective type of sentence. }
    \label{tab:summary_example}
    \begin{tabular}{|p{1.8in}|p{1.5in}|p{1.8in}|}
    \hline
    \textbf{Reference summary 
    (case name: 1994canlii3880(2,349 tokens in the original case decision))} & \textbf{Baseline GPT-4 } & \textbf{Arg. Segmentation Enhanced GPT-3 } \\
    \hline
    The accused, a young offender, pleaded guilty to 16 counts of auto theft.\colorbox{pink}{<Issue>} He was sentenced to 6 months open custody and 2 years probation.\colorbox{pink}{</Issue>} Prior to being sentenced, he spent 13 days on remand and 2 months under house arrest under s.7.1 of the Young Offenders Act. \colorbox{pink}{<Issue>}The Crown appealed the sentence.\colorbox{pink}{</Issue>} \colorbox{lime}{<Conclusion>}HELD: Appeal allowed.\colorbox{lime}{</Conclusion>} \colorbox{cyan}{<Reason>} 1)The accused was 15 years of age and had no record.\colorbox{cyan}{</Reason>} \colorbox{cyan}{<Reason>}He assisted the police once apprehended.\colorbox{cyan}{</Reason>} \colorbox{cyan}{<Reason>} 2)Although another young offender involved in these offences had received a sentence of 9 months open custody, uniformity of sentence was not a factor to be given great weight in comparing dispositions of young offenders.\colorbox{cyan}{</Reason>} \colorbox{lime}{<Conclusion>} 3)The sentence was increased to 10 months open custody.\colorbox{lime}{</Conclusion>}

    &This case involves a young offender, J.M., who pleaded guilty to multiple counts of auto theft, possession of stolen auto, and dangerous driving. The court considered several factors, including the extent of J.M.'s cooperation with the police, his father's involvement in his life, and the need for uniformity in sentencing among young offenders involved in similar crimes. Ultimately, the court decided that a fit disposition for J.M. would be four months open custody for the offenses prior to June 21, 1994, and six months open custody for the offenses committed on June 21, 1994, resulting in a total of ten months open custody and 18 months probation.
    
    &J.M. pled guilty to 16 counts of auto theft, one count of possession of stolen auto and one count of dangerous driving. He was sentenced to six months' open custody and two years' probation. The judge considered the police chase, deliberate damage to vehicles, and the large number of charges as aggravating factors. He also took into account J.M.'s age, lack of criminal record, cooperation with the police, and his father's responsible approach throughout the event. No, it is not possible to equate the two. Uniformity of sentencing is not a factor to be given great weight when comparing dispositions of young offenders, but an exception must be made in the case of youths who form part of the same enterprise and who are of the same age and circumstances. Probation does not sufficiently recognize the multiple nature of the offences and the malicious damage caused, but imposing the same disposition as [T.D.] would not be appropriate for several reasons. The appeal is allowed and the dispositions for the auto offences prior to June 21, 1994 will be set aside and replaced with four months open custody and 18 months probation, running consecutively for a total of ten months open custody. The terms of the probation order remain the same.\\ 

    \hline
  \end{tabular}
\end{table*}

\begin{table*}[h!]
    \caption{Argumentative segmentation enhanced GPT-3.5 generated summaries and baseline GPT-4 generated summaries versus the reference summary.}
    \label{tab:summary_example_long}
    \begin{tabular}{|p{1.8in}|p{1.5in}|p{1.8in}|}
    \hline

    \textbf{Reference summary 
    (case name: 1995canlii4465 (5,221 tokens in the original case decision))} & \textbf{Baseline GPT-4 } & \textbf{Arg. Segmentation Enhanced GPT-3.5 } \\
    \hline
    
  The plaintiffs statement of claim was served on the defendant in Ontario two days after it was issued.Two weeks later the prothonotary issued default judgment for \$8,000, plus \$2,880 in interest and \$404.42 in costs. \colorbox{pink}{<Issue>}The defendant applied, pursuant to Civil Procedure Rule 12.06, to set aside the default judgment, on the ground that the plaintiff did not meet the requirements for obtaining a default judgment from the prothonotary or, in the alternative, on the merits that it had, by affidavit evidence, raised triable issues and provided a reasonable excuse for its failure to file a defence.\colorbox{pink}{</Issue>} \colorbox{lime}{<Conclusion>}Dismissing the application with costs of \$1,000 to the plaintiff, \colorbox{lime}{</Conclusion>} \colorbox{cyan}{<Reason>}that although the plaintiff ought to have sought the court's leave to take default judgment such a procedural ommission did not, in the circumstances, prejudice the defendant. \colorbox{cyan}{</Reason>} \colorbox{cyan}{<Reason>}Further, while there were triable issues between the parties, the defendant's failure to file a defence was neither a mistake nor an oversight, but rather a deliberate decision so as not to attorn to Nova Scotia jurisdiction, and therefore was not a reasonable excuse.\colorbox{cyan}{</Reason>}

  &The defendant applied to set aside a default judgment obtained by the plaintiff, arguing procedural irregularities and that there were legitimate triable issues between the parties. The court dismissed the defendant's application, finding that any irregularities did not materially affect the defendant and that the cases relied upon by the defendant were not applicable to the circumstances of this case. The court also found that the defendant had failed to establish any reasonable excuse for its failure to file a defense, as it was a deliberate decision not to file a defense rather than a mistake or oversight.
    & The defendant argued three points in support of setting aside the default judgment: 1) procedural, 2) irregularity as of right, and 3) legitimate triable issues with reasonable excuse for failure to file a defence. The defendant's application to set aside the default judgment was dismissed because any irregularities that may have prompted the issuance of the Prothonotary's default judgment were not enough to make the Order a nullity, and the procedural omission did not affect the defendant. The appellant must show by affidavit facts which would indicate clearly that he had good defence to the action on the merits, and why his defence was not filed and delivered within the time limited by the Rules. Yes, I agree with Mr. Stobie that this was not a mistake or an oversight, but rather a deliberate decision not to file a defence. Ms. Roberts was aware that not filing a defence would result in default, and she had not provided reasonable excuse for the delay. The defendant's refusal to commit to defending in Nova Scotia or bring a motion to stay on the basis of forum non conveniens is not a reasonable excuse, and therefore the court will not exercise its discretion in favour of the defendant. The defendant's application is dismissed with costs to the plaintiff in the amount of One Thousand Dollars (\$1,000.00).
    \\
    \hline

    \hline
  \end{tabular}
\end{table*}


Since GPT-3.5 imposes the token request limitation, any input text longer than the limit should be chunked before submitting to the model. In our test dataset, almost half of the cases exceed the token limitation. For these longer opinions, segmenting them using our implementation of argument zoning would seem to be a reasonable step, possibly increasing the likelihood that GPT-3.5's summaries would include useful argument-related information. Table \ref{tab:summary_example_long} shows an example of generated summaries when the original case decision substantially exceeds GPT-3.5's token limit. As a result, we need to shorten the document first before feeding it to the model. Meanwhile, GPT-4 can handle the length of the original case decision.  We noticed that the baseline GPT-4 summary lacks some necessary details as compared to the argumentative segmentation enhanced approach. The latter included a more detailed presentation of the issue and conclusion and more of the reasons. The result was expected, since the input was shortened for the baseline. Despite the richness of information  that a GPT-3.5 summary provides, GPT-4  generates smoother summaries. The main reason is that GPT-4 has a longer context span than GPT-3.5. 

In terms of cost, we consider the current pricing scheme for both GPT-3.5 and GPT-4 based on the number of tokens
submitted to and generated by the model. The pricing of GPT-3.5 is set to \$0.02 per 1,000 tokens in both prompt and completion, while the pricing for GPT-4 is set to \$0.03 per 1,000 tokens in prompt and \$0.06 per 1,000 tokens in completion. The cost of using GPT-3.5 with argumentative segmentation to generate a summary is approximately \$0.19 on average. In comparison, the average cost for using GPT-4 is about \$1.31. This means that GPT-4 is approximately 10 times more expensive than GPT-3.5 for the summarization task. 

We also examined some of the summaries generated by the non-GPT models. The quality of summaries is clearly lower than GPT generated summaries. One possible reason is that large language models are trained on a much larger corpus and have more extensive model architectures, which makes them better few-shot or even zero-shot learners \cite{brown2020language}. 

\section{Limitations}
In this study, we focus on the effect of using argumentative segmentation on legal summarization. While we observed improvements in the model performance of legal summarization with argumentative segmentation, we also some coherency issues in the generated summaries. For example, ``Yes, I agree with Mr. Stobie" interrupt the information flow of the summary from Table \ref{tab:summary_example_long}. Thus, a systematic human evaluation of generated summaries is needed to further examine the performance of the models and address these coherency issues. 

Furthermore, reproducing our results may be challenging due to the proprietary nature of the OpenAI GPT models used in our experiments. Especially, we employed different combinations of control parameters in the experiment will further decrease the possibility of reproduction. Additionally, any updates or changes to the GPT models by OpenAI may result in changes to performance and results. So it is crucial to develop methods to increase the reproducibility of the results.
\section{Conclusion and Future Work}
We have proposed a novel task of extracting argumentative segments that include the main points of legal case decisions.  We further proposed to utilize these argumentative segments to guide a summarizer. Our experiments with GPT-3.5, GPT-4 and other models showed that the argumentative segmentation enhanced method can improve the automatic evaluation scores of generated summaries. This method also overcomes the request token limitation imposed by GPT-3.5. Our findings reveal a boost in performance across all types of automatic evaluations scores using the predicted argumentative segments. Additionally, we observed that GPT-4 tends to produce more coherent summaries compared to GPT-3.5. 

For future work, we will further explore methods to ensure more reliable performance of the proprietary models. Furthermore, we plan to investigate alternative prompt engineering techniques for the summarization task. Due to the nature of generative models, a systematic human evaluation on the generated summaries are much needed in the future.

\begin{acknowledgments}
This work has been supported by grants from the Autonomy through Cyberjustice Technologies Research Partnership at the University of Montreal Cyberjustice Laboratory and
the National Science Foundation, grant no. 2040490, FAI: Using AI to Increase Fairness by Improving Access to Justice. The Canadian Legal Information Institute provided
the corpus of paired legal cases and summaries. This work was supported in part by the University of Pittsburgh Center for Research Computing through the resources provided. Specifically, this work used the H2P cluster, which is supported by NSF award number OAC-2117681.
\end{acknowledgments}

\bibliography{sample-ceur}

\appendix

\end{document}